\documentclass{article}

\usepackage[preprint]{neurips_2025}

\usepackage[utf8]{inputenc} 
\usepackage[T1]{fontenc}    
\usepackage{hyperref}       
\usepackage{url}            
\usepackage{booktabs}       
\usepackage{amsfonts}       
\usepackage{nicefrac}       
\usepackage{microtype}      
\usepackage{xcolor}         
\usepackage{graphicx}
\usepackage{amsmath}
\usepackage{multirow}
\definecolor{lightgreen}{RGB}{144, 238, 144}
\definecolor{deepgreen}{RGB}{0,136,55}
\newcommand{\rng}[1]{{\tiny$\pm$#1}}
\usepackage{algorithm}
\usepackage{algpseudocode}
\usepackage{multirow}
\usepackage{graphicx}
\usepackage{subfig}
\usepackage{adjustbox}
\title{HAD: Hybrid Architecture Distillation Outperforms Teacher in Genomic Sequence Modeling}
\author{
    \textbf{Hexiong Yang}$^{1,3}$ \textbf{Mingrui Chen}$^{2,3,4}$\\
    \textbf{Huaibo Huang}$^{2,3}$ \textbf{Junxian Duan}$^{2,3}$ \textbf{Jie Cao}$^{2,3}$ \textbf{Zhen Zhou}$^{5}$ \textbf{Ran He}$^{2,3}$ \\
    $^1$School of Advanced Interdisciplinary Science, University of Chinese Academy of Sciences\\
    $^2$School of Artificial Intelligence, University of Chinese Academy of Sciences\\
    $^3$NLPR\&MAIS, Institute of Automation, Chinese Academy of Sciences\\
    $^4$Zhongguancun Academy $^5$Nanjing University \\
    \texttt{yanghexiong2002@gmail.com}, \texttt{charmier2003@gmail.com}, \texttt{rhe@nlpr.ia.ac.cn}
}
\begin{document}
\maketitle
\begin{abstract}
Inspired by the great success of Masked Language Modeling (MLM) in the natural language domain, the paradigm of self-supervised pre-training and fine-tuning has also achieved remarkable progress in the field of DNA sequence modeling.
However, previous methods often relied on massive pre-training data or large-scale base models with huge parameters, imposing a significant computational burden.
To address this, many works attempted to use more compact models to achieve similar outcomes but still fell short by a considerable margin. In this work, we propose a \underline{H}ybrid \underline{A}rchitecture \underline{D}istillation (HAD) approach, leveraging both distillation and reconstruction tasks for more efficient and effective pre-training.
Specifically, we employ the NTv2-500M as the teacher model and devise a grouping masking strategy to align the feature embeddings of visible tokens while concurrently reconstructing the invisible tokens during MLM pre-training.
To validate the effectiveness of our proposed method, we conducted comprehensive experiments on the Nucleotide Transformer Benchmark and Genomic Benchmark. Compared to models with similar parameters, our model achieved excellent performance. \textit{\textbf{More surprisingly}}, it even surpassed the \textit{distillation ceiling}-teacher model on some sub-tasks, which is more than \textbf{\textit{500 $\times$}} larger. Lastly, we utilize t-SNE for more intuitive visualization, which shows that our model can gain a sophisticated understanding of the intrinsic representation pattern in genomic sequences.
\end{abstract}
\section{Introduction}
\label{Sec_Introduction}

In the realm of DNA sequence modeling, the paradigm of self-supervised pre-training followed by fine-tuning is catalyzing significant advancements, fundamentally reshaping how genomic data is interpreted and utilized \cite{chen2022sequence, zhang2024artificial}. Within this transformative landscape, masked language modeling (MLM) has prominently emerged as a primary technique. By pre-training models on vast, unlabeled DNA datasets, these methods can learn effective representations for downstream genomic tasks \cite{ji2021dnabert,zhou2023dnabert,dalla2024nucleotide,nguyen2023hyenadna,schiff2024caduceus,DBLP:conf/icml/LiW000Z0L24}. 
These learned representations are foundational for numerous downstream genomic tasks, greatly improving predictive capabilities and fostering deeper biological insights \cite{linder2025predicting, DBLP:conf/icml/LiW000Z0L24}.

Scaling up the parameter size of models, especially in Transformers-based work~\cite{ji2021dnabert, zhou2023dnabert, dalla2024nucleotide}, is a prevalent method for enhancing pre-trained models. Though this approach can often bring certain performance gain, it inevitably results in considerably higher computational demands \cite{ru2025we}. 

Consequently, an alternative and also crucial research direction that focuses on novel, compact and efficient architectures has emerged~\cite{nguyen2023hyenadna, schiff2024caduceus, consens2025genomic}.
However, although these compact architectures provide desirable computational efficiency, they often struggle to match the in-depth representation learning and performance of those larger or more extensively pre-trained counterparts~\cite{dalla2024nucleotide, schiff2024caduceus}.

Compact models, despite their efficient pre-training capable of processing tens of billions of nucleotide tokens \cite{schiff2024caduceus}, often hit capacity limits early, restricting their ability to learning complex patterns from massive genomic data. Conversely, larger models undergo far more extensive training~\cite{dalla2024nucleotide} for deeper extraction of subtle biological feature. Thus, achieving profound representation learning in compact models remains a key challenge.

To overcome these limitations, we propose a novel framework for genomic sequence modeling with \underline{H}ybrid \underline{A}rchitecture \underline{D}istillation (HAD). HAD uses a hybrid student architecture to capture a wide range of DNA sequence features, from key local feature to the global interactions, within only \textbf{1M} parameter. 
Based on a bidirectional Gated Delta Net (GDN) \cite{yang2024gated}, it combines linear complexity with adaptive memory control via two complementary mechanisms: the gating mechanism selectively erases irrelevant or redundant non-functional sequence segments, the delta update rule accurately modifies memory by identifying specific short sequences. 
To integrate comprehensive global information, this GDN backbone is augmented with a self-attention layer \cite{dao2022flashattention,dao2023flashattention2}. This hybrid approach harnesses combined strengths, effectively integrating GDN's proficiency in capturing local and long-range sequential patterns with attention's capacity for unifying global context.

Our model empowers compact architecture with deep biological understanding through hybrid learning tasks, implemented within an innovative parallel dual-branch pretraining framework.
These tasks distinctively combine two complementary objectives: a high-level feature alignment with a large teacher model using \textit{visible} DNA nucleotides, and a low-level nucleotide reconstruction task focusing on \textit{masked} positions. 
For the high-level alignment, our devised grouping masking strategy directs the student to align its feature embeddings of visible tokens with those from the teacher model, Nucleotide Transformer v2~\cite{dalla2024nucleotide}(>500M), to gain more sophisticated biological insights.
Concurrently, the low-level reconstruction branch try to predict the original identities of masked nucleotides by using the learned representation of visible nucleotides from aliment branch as context, which is implemented by a cross-attention mechanism and motivates our model to learn fundamental DNA sequence patterns and local grammar. 
This hybrid framework ensures HAD develops both profound representation and fine-grained understanding of DNA sequence.

To validate our proposed method's effectiveness, we conduct comprehensive evaluation on the widely used Nucleotide Transformer Benchmark and Genomic Benchmark. Our compact model with only $1$M-parameter exhibited notable efficacy, not only outperforming competing models of similar parameter size but also \textit{\textbf{surprisingly}} outperforms its teacher model with $500$M-parameter.
Furthermore, we perform more intuitive visual analysis based on t-SNE, which further reveals that our model can learn intrinsic feature pattern and discriminative genomic representation from diverse DNA categories.
These observations confirm our approach's effectiveness for genomic sequence modeling and various downstream tasks.
\section{Related Work}
\label{Sec_Related_Work}

\subsection{Network Architecture For DNA Modeling}

In recent years, research in DNA sequence modeling has increasingly focused on the development of more efficient model architectures, particularly in the context of Transformer-based models. Notable works such as DNABERT \cite{ji2021dnabert}, DNABERT2 \cite{zhou2023dnabert}, and Nucleotide Transformer \cite{dalla2024nucleotide} have successfully employed standard Transformers as their backbone networks, achieving impressive performance in genomic sequence tasks. However, these models are not without limitations, especially their scalability to long-sequence modeling and their relatively high inference costs.
To address these challenges, recent advancements have turned to more efficient modeling approaches\cite{DBLP:journals/corr/abs-2411-07635,DBLP:conf/cvpr/FanHCL024,DBLP:journals/corr/abs-2303-17803,fan2024vision,fan2024semantic,gu2023mamba,dao2024transformers,yang2024parallelizing,yang2024gated,behrouz2024titans,sun2023retentive}. For instance, HyenaDNA \cite{nguyen2023hyenadna} introduces the Hyena operator, which reduces the model size to approximately 6.6M parameters while extending the model's capacity to handle sequences up to 1M in length. Similarly, Caduceus \cite{schiff2024caduceus} proposes a bidirectional and RC-equivariant Mamba block as the backbone, successfully incorporating the concept of selective structured state space models (SSMs)\cite{gu2023mamba,dao2024transformers} into the domain of DNA sequence modeling.
In the realm of RNN-based models, recent studies have enhanced the global modeling capacity by incorporating a limited number of attention layers into the architecture. Additionally, recent works have introduced novel computational strategies, such as the Delta Rule \cite{yang2024parallelizing,yang2024gated} and the Titans \cite{behrouz2024titans}, which aim to improve memory management and retrieval performance for sequence modeling tasks\cite{arora2023zoology,wen2024rnns,yin2025atri,akyurek2024context}.
These developments indicate that adopting more efficient architectural designs and advanced computational strategies can overcome the inherent limitations of existing models, offering promising avenues for progress in DNA sequence modeling.
This paper explores the potential of hybrid architectures as backbone networks for DNA modeling, aiming to advance the field through these innovative approaches.

\subsection{Knowledge Distillation For DNA Modeling}
Knowledge distillation (KD) \cite{gou2021knowledge,tang2020understanding,busbridge2025distillation} is a model compression technique that transfers knowledge from a large, high-capacity teacher model to a lightweight student model, enabling the latter to mimic the teacher’s behavior while reducing computational costs. Initially proposed by \cite{hinton2015distilling}, KD leverages soft targets derived from the teacher’s output distribution rather than relying solely on ground-truth labels, thereby capturing richer inter-class relationships and enhancing generalization. 
Over time, KD has evolved into diverse paradigms, including feature-based distillation (\textit{e.g.}, aligning intermediate representations), contrastive distillation (\textit{e.g.}, preserving sample similarity structures), and relational distillation (\textit{e.g.}, modeling geometric relationships). Recent advancements extend KD to cross-architecture settings, enabling knowledge transfer between heterogeneous model families (\textit{e.g.}, Transformer→MLP), and self-distillation frameworks where the student iteratively refines its own outputs.
In masked image modeling (MIM)\cite{cao2020parametric,dong2023peco,xie2022simmim,woo2023convnext,oquab2023dinov2,zhou2022image}, where models learn by reconstructing masked regions of images, distillation has been instrumental in compressing large vision transformers (ViTs). For instance, feature-based distillation aligns intermediate attention maps between teacher and student models, preserving spatial-semantic patterns critical for reconstruction. While KD in DNA pretraining remains underexplored, insights from related domains suggest promising directions, such as Distilled DeepConsensus \cite{belyaeva2022knowledge} and FinDNA \cite{yu2025self}, which applied KD and self-distillation techniques in the DNA correction and prediction tasks respectively. Analogously, DNA pretraining could leverage feature distillation to align latent representations of genomic sequences between teacher and student models, preserving motifs and regulatory patterns.
In this paper, we will dive into this question and explore the potential of the knowledge distillation for hybrid architectures.
\section{Methodology}
\label{Sec_Methodology}

\subsection{Overall Pipeline}
\label{subsec:Overall_Pipeline}
\begin{figure}[tbp]
    \centering
    \vspace{-0.1cm}
    \includegraphics[width=\linewidth]{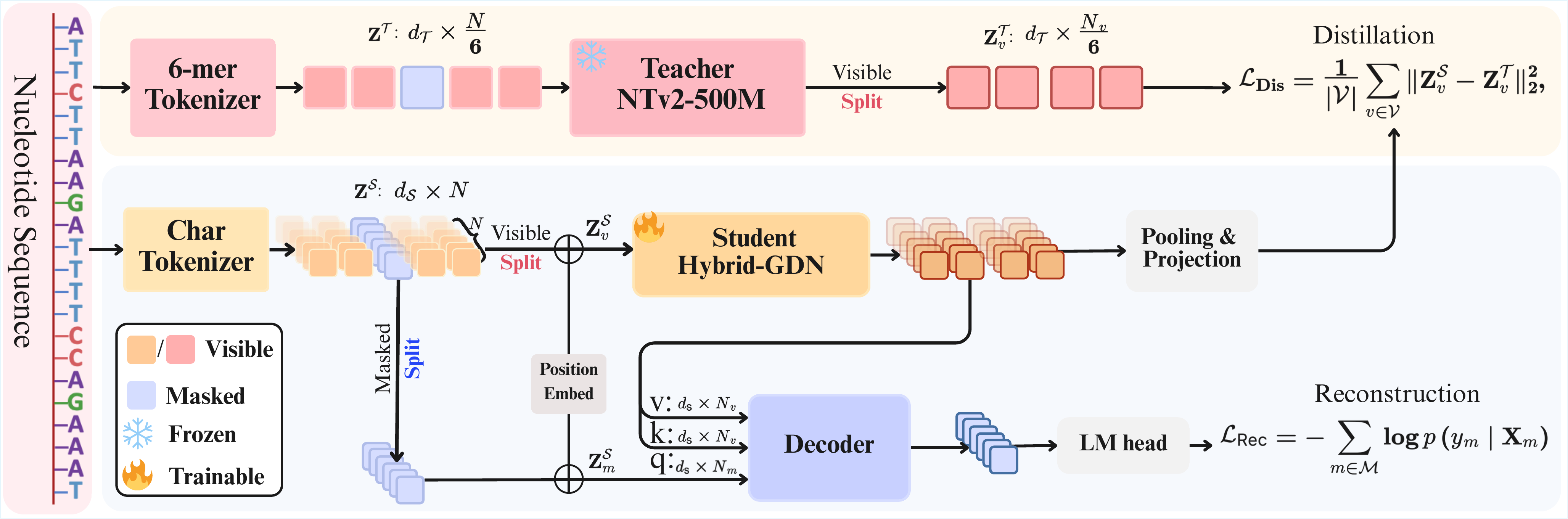}
    \caption{Proposed \textbf{H}ybrid \textbf{A}rchitecture \textbf{D}istillation ($\textbf{HAD}$) pre-training pipeline. The upper branch performs feature alignment on visible nucleotides, distilling \textit{high-level} knowledge from a pre-trained teacher model to the student model. The lower branch focuses on the \textit{low-level} reconstruction of masked nucleotides, leveraging contextual information from the student's visible nucleotide representations.}
    \label{fig:HAD}
    \vspace{-0.1cm}
\end{figure}

Traditional self-supervised learning for DNA sequences, such as Masked Language Modeling (MLM), typically processes a partially masked input sequence $\mathbf{X}_{m}$ (derived from $\mathbf{X}$) through a unified encoder to predict the original nucleotides at masked positions ${m}$, optimized via a reconstruction objective:

\begin{equation}
\mathcal{L}_{\text{Rec}} = - \sum_{{m} \in \mathcal{M}} \log p\left(y_{m} \mid \mathbf{X}_{m}\right) ,
\label{eq:rec_loss}
\end{equation}

\noindent where $\mathcal{M}$ is the set of masked positions and $p\left(y_{m} \mid \mathbf{X}_{m}\right)$ represents the predicted probability of $y_i$ given $\mathbf{X}_{\mathcal{M}}$. However, conventional MLM may not fully enable compact models to learn the deep features seen in much larger, extensively pre-trained models, especially when leveraging massive datasets. Thus, our Hybrid Architecture Distillation (HAD) framework significantly innovates upon this to bridge this gap by introducing a dual-branch pipeline. This design enables synergistic learning through two distinct yet complementary objectives: high-level feature alignment on visible nucleotides and low-level  reconstruction of masked nucleotides, moving beyond the single-stream processing of conventional MLM.

The overall pipeline of HAD is illustrated in Figure~\ref{fig:HAD}. It begins by conceptually dividing the input sequence $\mathbf{X}$ into visible nucleotides $\mathbf{X}_{v}$ and masked positions $\mathbf{X}_{m}$. In the first branch, the student model $\mathcal{S}$ processes $\mathbf{X}_{v}$ (via its character-level tokenizer) into a hidden representation $\mathbf{Z}^{\mathcal{S}}_{v}$. This is then aligned with the corresponding visible representation $\mathbf{Z}^{\mathcal{T}}_{v}$ derived from a large, pre-trained teacher model $\mathcal{T}$ (which processes the full $\mathbf{X}$ using its \text{k-mer} tokenizer and backbone, followed by filtering for visible parts). This feature-level distillation provides explicit high-level guidance. The second branch reconstructs the nucleotides at masked positions. For this, a decoder module integrates contextual information from the student's visible nucleotide representations $\mathbf{Z}^{\mathcal{S}}_{v}$ with initial embeddings derived from the masked positions $\mathbf{X}_{\mathcal{M}}$. This integration yields context-aware representations for the masked nucleotides, $\mathbf{Z}^{\mathcal{S}}_{\mathcal{M}}$. An LM Head subsequently maps these representations $\mathbf{Z}^{\mathcal{S}}_{m\mathcal{M}}$ to vocabulary logits to predict their original types, optimized with a loss analogous to $\mathcal{L}_{\text{Rec}}$. Thus, HAD distinctively conditions masked nucleotide reconstruction on information from the visible pathway, a key departure from standard MLM's reliance on local masked context alone.

\subsection{Hybrid Learning Tasks}
\paragraph{Masking Strategy for Tokenizer Mismatch.}
Traditional random nucleotide masking is unsuitable for our hybrid learning task, as such strategies can create information inconsistencies and leakage during feature alignment between the $k$-mer level teacher and character-level student models, impairing distillation. To ensure consistent information and effective feature alignment, we therefore propose a two-stage mask sampling method. 

\begin{figure}[t]
    \centering
    \includegraphics[width=\linewidth]{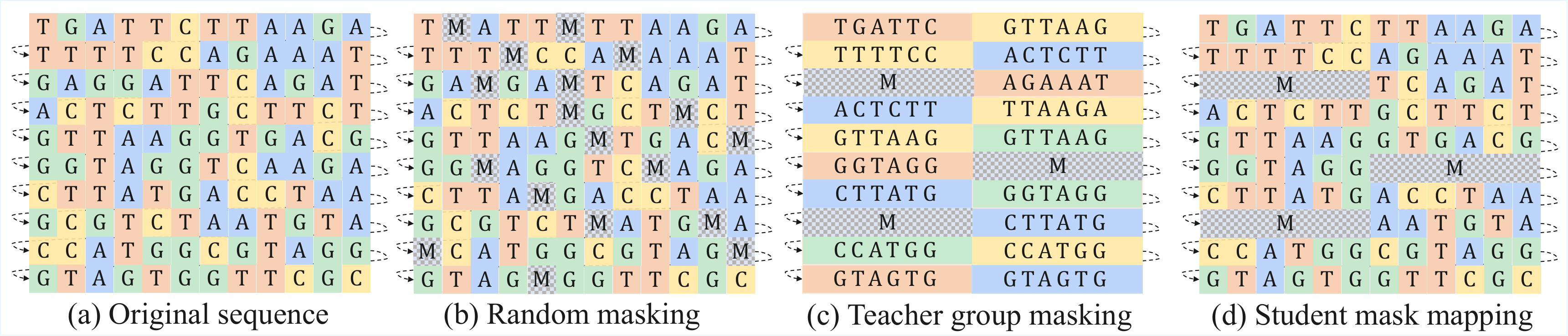}
    \vspace{-0.2cm}
    \caption{Two-stage masking strategy in HAD. This strategy is designed to prevent information leakage and enhance feature learning for the student model during distillation.}
    \label{fig:masking_strategy}
    \vspace{-0.2cm}
\end{figure}

Specifically, the first stage implements ``teacher group masking'' at the $k$-mer level (Figure~\ref{fig:masking_strategy}c). Here, we randomly select 15\% of $k$-mer units (\textit{e.g.}, for a sequence of length $N$ and 6-mers, $N/6$ units) to define the masked regions for the teacher model. 
By masking entire $k$-mer blocks, it presents a more structurally coherent and challenging masked context. For our student model, which operates at a character-level, this encourages learning from larger obscured spans during distillation, thereby fostering the acquisition of more valuable, high-level features. The second stage then involves mapping these $k$-mer level mask indices to the corresponding character-level positions for the student model (Figure~\ref{fig:masking_strategy}d). This two-stage approach ensures consistent mask positioning between the teacher and student, prevents information leakage, and importantly, enhances the quality of feature learning for the student through more meaningful group-level masking. 

\paragraph{Feature Alignment of Visible Nucleotides.}
With the visible and masked nucleotide positions established by our two-stage masking strategy, the first of our hybrid learning tasks focuses on feature alignment using these visible nucleotides. These visible segments are processed through the student model's hybrid architecture (detailed in Section~\ref{Subsec_Hybrid_Architecture}) to obtain its representation $\mathbf{Z}^{\mathcal{S}}_{v}$. This student representation is then aligned with the corresponding features $\mathbf{Z}^{\mathcal{T}}_{v}$ derived from a large-scale teacher model, NTv2 (a pure Transformer architecture with 500M parameters). Pre-trained on a multi-species dataset using over 1 trillion tokens, the NTv2 teacher model exhibits strong biological representation capabilities. The primary goal of this alignment is to enable the student model to inherit the teacher’s sophisticated biological expertise and learned high-level features from the visible portions of the sequence.

The alignment process faces two main challenges: differing hidden dimensions and sequence length. The student model has a hidden dimension of $d_\mathcal{S} = 128$, while the teacher model has $d_\mathcal{T} = 1024$. Additionally, the student model uses a character-level tokenizer, producing sequences of length $L = N$, whereas the teacher model uses a \text{k-mer} tokenizer with $k = 6$, resulting in sequences of length $L_{\text{k-mer}} = \frac{N}{6}$.
To address these, we first apply average pooling to the student model’s sequence representations, reducing the sequence length from $L$ to $L_{\text{k-mer}}$ to match the teacher model’s output. This is done over non-overlapping 6-mer windows. After aligning the sequence length, we use a projection layer to map the student model's hidden representations from $d_S$ to $d_T$. These two operations align both the sequence length and feature dimensions, facilitating appropriate feature alignment.

The feature alignment is achieved by minimizing the Mean Squared Error (MSE) loss between the student and teacher model representations. Since only visible nucleotides are aligned, the teacher model extracts representations for visible nucleotides based on pre-sampled mask indices. The MSE loss is computed as:
\begin{equation}
    \mathcal{L}_{\mathrm{Dis}} = \frac{1}{\left|\mathcal{V}\right|} \sum_{v \in \mathcal{V}} \| \mathbf{Z}^{\mathcal{S}}_{v} - \mathbf{Z}^{\mathcal{T}}_{v} \|_2^2,
\end{equation}
where $\mathcal{V}$ denotes the set of visible nucleotide positions. Within the sum, $\mathbf{Z}^{\mathcal{S}}_{v}$ is the student model’s representation at a visible position $v \in \mathcal{V}$, and $\mathbf{Z}^{\mathcal{T}}_{v}$ is the teacher model’s representation at the same position $v$. By minimizing this loss, we ensure that the student model effectively aligns its visible nucleotide representations with the teacher model’s.

\paragraph{Reconstruction of Masked Nucleotides.}
To preserve the model's capability for low-level nucleotide understanding, a masked nucleotide reconstruction task remains essential. However, due to our framework's clear division of visible and masked nucleotide processing pathways, a dedicated decoder mechanism is necessary for this reconstruction. In our approach for this task, the masked nucleotide positions are initialized randomly rather than using a fixed \texttt{[MASK]} token. 
We also ensure that positional information is added to the representations of both visible and masked nucleotides before they enter their respective model pathways. This explicit positional encoding preserves crucial spatial information, enabling accurate reconstruction of the original masked nucleotide positions.

For the Decoder, we use Cross Attention(CA), where the masked nucleotides act as the query, and the visible nucleotides serve as both the key and value. The masked nucleotides' representations, as queries, attend to the visible nucleotides' representations to produce the corresponding masked sequence. The CA operation is computed as follows:
\begin{equation}
    \operatorname{CA}\left(\mathbf{Q}_{m}, \mathbf{K}_{v}, \mathbf{V}_{v}\right)=\operatorname{softmax}\left(\frac{\mathbf{Q}_{m} \mathbf{K}_{v}^{T}}{\sqrt{d_{k}}}\right) \mathbf{V}_{v},
\end{equation}
where $\mathbf{Q}_m$ is the query (masked nucleotides), $\mathbf{K}_v$ and $\mathbf{V}_v$ are the key and value (visible nucleotides), and $d_k$ is the dimensionality of the keys. This attention mechanism allows the model to focus on the relevant information in the visible nucleotides while reconstructing the masked nucleotides.

Finally, an LM Head maps the resulting masked representations $\mathbf{Z}^{\mathcal{S}}_{m}$ to vocabulary logits, producing a probability distribution for each masked nucleotide. The optimization objective for this reconstruction is to minimize a cross-entropy function analogous to $\mathcal{L}_{\text{Rec}}$ (Same as Equation~\eqref{eq:rec_loss}).

\subsection{Hybrid Student Architecture}
\label{Subsec_Hybrid_Architecture}

\begin{figure}[tbp]
    \centering
    \includegraphics[width=\linewidth]{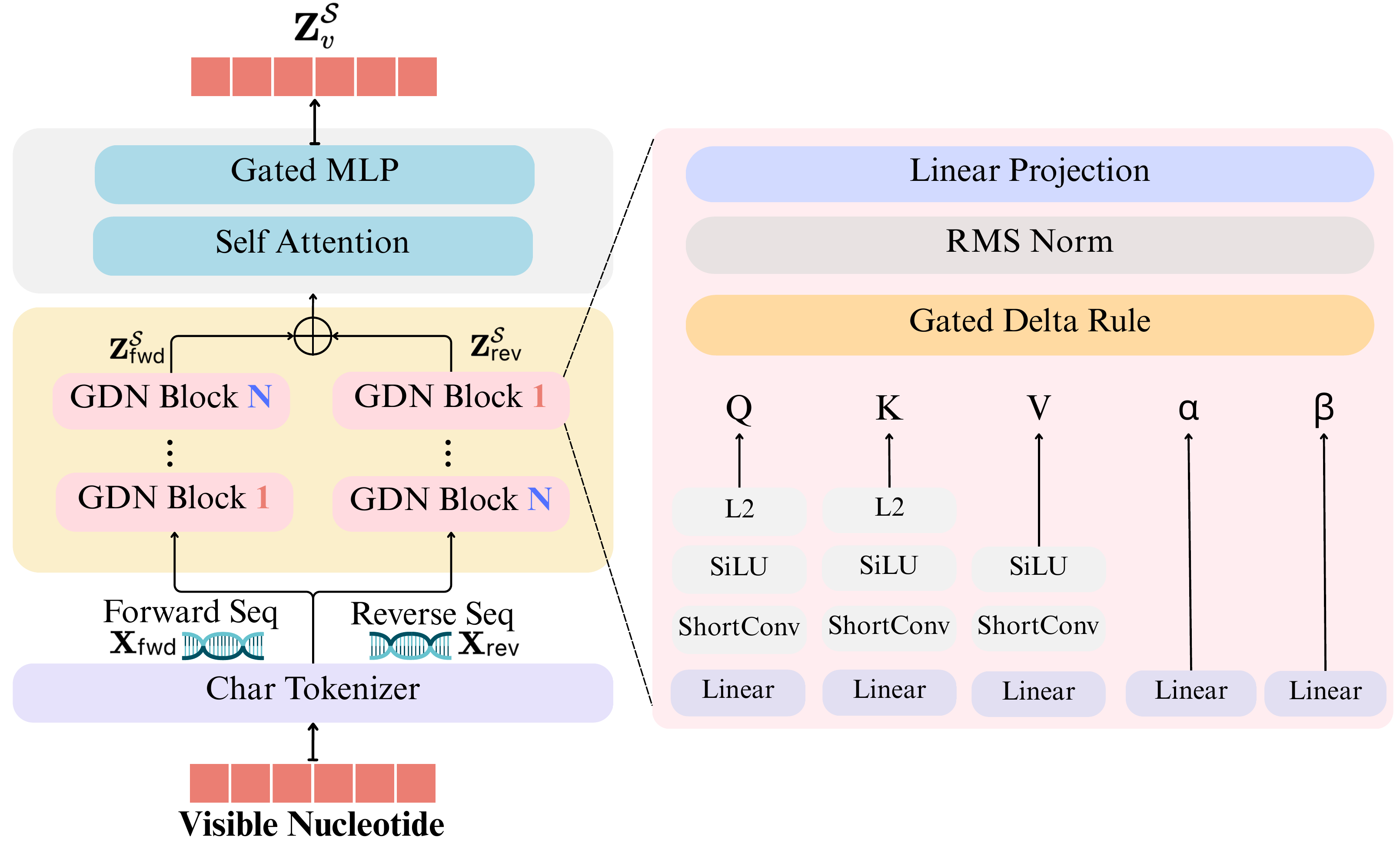}
    \caption{Hybrid architecture of our student model, combining a bidirectional Gated Delta Net (GDN) backbone with a self-attention layer for efficient sequential processing and global context integration within a compact 1.1M parameter budget. Within the GDN, $\alpha$ serves as a data-dependent gate controlling memory erasure, while $\beta$ acts as the update strength from the delta rule.}

    \label{fig:hybridGDN}
    \vspace{-0.2cm}
\end{figure}

Our hybrid architecture for DNA sequence modeling is founded on the Gated Delta Net (GDN) \cite{yang2024parallelizing,yang2024gated}. The state $\mathbf{S}_t$ of GDN at each time step $t$ is dynamically updated based on its previous state $\mathbf{S}_{t-1}$ and current inputs, following the principle:
\begin{equation}
\mathbf{S}_{t}=\mathbf{S}_{t-1}\left(\alpha_{t}\left(\mathbf{I}-\beta_{t} \boldsymbol{k}_{t} \boldsymbol{k}_{t}^{T}\right)\right)+\beta_{t} \boldsymbol{v}_{t} \boldsymbol{k}_{t}^{T}.
\label{eq:gdn_update}
\end{equation}

Equation~\eqref{eq:gdn_update} details how GDN dynamically updates its memory $\mathbf{S}_t$ for our DNA sequence modeling, using complementary ``Gated'' and ``Delta Rule'' mechanisms. The gating, via $\alpha_t$ and the $(\mathbf{I}-\beta_{t} \boldsymbol{k}_{t} \boldsymbol{k}_{t}^{T})$ term, allows the model to selectively clear or retain prior nucleotide context, effectively filtering information from less relevant DNA segments based on the current key $\boldsymbol{k}_t$. The Delta Rule component, $\beta_{t} \boldsymbol{v}_{t} \boldsymbol{k}_{t}^{T}$, then precisely incorporates features from the current input nucleotides (key $\boldsymbol{k}_t$, value $\boldsymbol{v}_t$), ensuring significant DNA patterns update the memory. 
To enable bidirectional modeling, we process the original sequence $\mathbf{X}_{\text{fwd}}$ and its reverse $\mathbf{X}_{\text{rev}}$ with separate GDN modules, yielding forward $\mathbf{Z}^{\mathcal{S}}_{\text{fwd}}$ and reverse $\mathbf{Z}^{\mathcal{S}}_{\text{rev}}$ hidden states. These are combined by reversing $\mathbf{Z}^{\mathcal{S}}_{\text{rev}}$ and adding it to $\mathbf{Z}^{\mathcal{S}}_{\text{fwd}}$, allowing the model to capture dependencies from both upstream and downstream contexts. To achieve efficient GPU utilization, GDN is parallelized using a chunk-wise method.

To further integrate global sequence information, the bidirectional GDN backbone is enhanced with an attention mechanism\cite{vaswani2017attention}. Specifically, following the final bidirectional GDN module, we append a single self-attention layer, implemented using Flash Attention \cite{dao2022flashattention,dao2023flashattention2}. The output from this attention layer is then processed by a simple Gated MLP. This completes our hybrid architecture, designed to balance efficient sequential modeling with global contextual understanding.

\section{Experiments}
\label{Sec_Experiments}
In this section, we present the experimental setup and results for evaluating our proposed method.
Our model is evaluated against state-of-the-art baselines on the Nucleotide Transformer and Genomic Benchmarks\cite{grevsova2023genomic}. 
We also use t-SNE for visualization, aiming to validate the feature learning capability transferred from the teacher to the student model within the HAD framework.

\subsection{Experiments Setting}
\begin{table}[tbp]
\centering
\caption{
    \textbf{Nucleotide Transformer Benchmark Results.} Performance of HAD against baseline models and its NTv2 teacher. 
    Results are means from 10-fold cross-validation with 10 random seeds; best performance is in \textbf{bold}, second-best is \underline{underlined}. 
    Error bars represent the range (maximum-minimum) across the 10 seeds. 
    The final column, $\Delta_{\text{Student-Teacher}}$, shows the performance difference between HAD and its teacher.
}
\label{tabs:NT_result}
\scriptsize
\makebox[\linewidth][c]{
\resizebox{1.02\linewidth}{!}{
\begin{tabular}{lccccccc}
\toprule
\textbf{Dataset} & \textbf{Enformer}\cite{avsec2021effective} & \textbf{DNABERT-2}\cite{zhou2023dnabert} & \textbf{HyenaDNA}\cite{nguyen2023hyenadna} & \textbf{Caduceus}\cite{schiff2024caduceus} & \textbf{NT(Teacher)}\cite{dalla2024nucleotide} & \textbf{HAD(Student)} & $\Delta_{\textbf{Student-Teacher}}$ \\
\textbf{Param.} & 252M & 117M & 1.6M & 1.9M & 498.3M & 1.1M & \textcolor{deepgreen}
{\textbf{-497.2M}}\\
\midrule
\multicolumn{7}{l}{ \textit{\textbf{Histone Markers}}} \\
H3       & 0.719 \rng{0.048} & 0.785 \rng{0.033} & 0.779 \rng{0.037} & \underline{0.815} \rng{0.048} & 0.784 \rng{0.047} & \textbf{0.822} \rng{0.007} & \textcolor{deepgreen}{\textbf{+3.8\%}}\\
H3K14ac  & 0.288 \rng{0.077} & 0.591 \rng{0.028} & 0.612 \rng{0.065} & \underline{0.631} \rng{0.026} & 0.551 \rng{0.021} & \textbf{0.684} \rng{0.041} & \textcolor{deepgreen}{\textbf{+13.6\%}}\\
H3K36me3 & 0.344 \rng{0.055} & 0.591 \rng{0.020} & 0.613 \rng{0.041} & 0.601 \rng{0.129} & \underline{0.625} \rng{0.013} & \textbf{0.653} \rng{0.031} & \textcolor{deepgreen}{\textbf{+2.8\%}}\\
H3K4me1  & 0.291 \rng{0.061} & 0.511 \rng{0.028} & 0.512 \rng{0.024} & 0.523 \rng{0.039} & \underline{0.550} \rng{0.021} & \textbf{0.571} \rng{0.037} & \textcolor{deepgreen}{\textbf{+2.1\%}}\\
H3K4me2  & 0.211 \rng{0.069} & 0.336 \rng{0.040} & 0.455 \rng{0.095} & \underline{0.487} \rng{0.170} & 0.319 \rng{0.045} & \textbf{0.562} \rng{0.033} & \textcolor{deepgreen}{\textbf{+24.3\%}}\\
H3K4me3  & 0.158 \rng{0.072} & 0.352 \rng{0.077} & \underline{0.549} \rng{0.056} & 0.544 \rng{0.045} & 0.410 \rng{0.033} & \textbf{0.643} \rng{0.052} & \textcolor{deepgreen}{\textbf{+23.3\%}}\\
H3K79me3 & 0.496 \rng{0.042} & 0.613 \rng{0.030} & 0.672 \rng{0.048} & \underline{0.697} \rng{0.077} & 0.626 \rng{0.026} & \textbf{0.712} \rng{0.022} & \textcolor{deepgreen}{\textbf{+8.6\%}}\\
H3K9ac   & 0.420 \rng{0.063} & 0.542 \rng{0.029} & 0.581 \rng{0.061} & \underline{0.622} \rng{0.030} & 0.562 \rng{0.040} & \textbf{0.656} \rng{0.023} & \textcolor{deepgreen}{\textbf{+9.4\%}}\\
H4       & 0.732 \rng{0.076} & 0.796 \rng{0.027} & 0.763 \rng{0.044} & \textbf{0.811} \rng{0.022} & 0.799 \rng{0.025} & \underline{0.806} \rng{0.018} & \textcolor{deepgreen}{\textbf{+0.7\%}}\\
H4ac     & 0.273 \rng{0.063} & 0.463 \rng{0.041} & 0.564 \rng{0.038} & \underline{0.621} \rng{0.054} & 0.495 \rng{0.032} & \textbf{0.654} \rng{0.039} & \textcolor{deepgreen}{\textbf{+15.9\%}}\\
\midrule
\multicolumn{6}{l}{\textit{\textbf{Enhancer Annotation}}}\\
Enhancer & 0.451 \rng{0.108} & 0.516 \rng{0.098} & 0.517 \rng{0.117} & 0.546 \rng{0.073} & \underline{0.548} \rng{0.144} & \textbf{0.571} \rng{0.075} & \textcolor{deepgreen}{\textbf{+2.3\%}}\\
Types    & 0.309 \rng{0.134} & 0.423 \rng{0.051} & 0.386 \rng{0.185} & \underline{0.439} \rng{0.054} & 0.424 \rng{0.132} & \textbf{0.467} \rng{0.073} & \textcolor{deepgreen}{\textbf{+4.3\%}}\\
\midrule
\multicolumn{7}{l}{\textit{\textbf{Promoter Annotation}}}\\
All      & 0.954 \rng{0.006} & \underline{0.971} \rng{0.006} & 0.960 \rng{0.005} & 0.970 \rng{0.004} & \textbf{0.976} \rng{0.006} & 0.968 \rng{0.004} & \textcolor{gray}{\textbf{-0.8\%}}\\
Non-TATA & 0.955 \rng{0.010} & \underline{0.972} \rng{0.005} & 0.959 \rng{0.008} & 0.969 \rng{0.011} & \textbf{0.976} \rng{0.005} & 0.968 \rng{0.005} & \textcolor{gray}{\textbf{-0.8\%}}\\
TATA     & \underline{0.960} \rng{0.023} & 0.955 \rng{0.021} & 0.944 \rng{0.040} & 0.953 \rng{0.016} & \textbf{0.966} \rng{0.013} & 0.958 \rng{0.009} & \textcolor{gray}{\textbf{-0.8\%}}\\
\midrule
\multicolumn{7}{l}{\textit{\textbf{Splice Site Annotation}}}\\
All      & 0.848 \rng{0.019} & 0.939 \rng{0.009} & \underline{0.956} \rng{0.011} & 0.940 \rng{0.027} & \textbf{0.983} \rng{0.008} & 0.911 \rng{0.016} & \textcolor{gray}{\textbf{-7.2\%}}\\
Acceptor & 0.914 \rng{0.028} & \underline{0.975} \rng{0.006} & 0.958 \rng{0.010} & 0.937 \rng{0.033} & \textbf{0.981} \rng{0.011} & 0.858 \rng{0.016} & \textcolor{gray}{\textbf{-12.3\%}}\\
Donor    & 0.906 \rng{0.027} & \underline{0.963} \rng{0.006} & 0.949 \rng{0.024} & 0.948 \rng{0.025} & \textbf{0.985} \rng{0.022} & 0.887 \rng{0.045} & \textcolor{gray}{\textbf{-9.8\%}}\\
\bottomrule
\end{tabular}
}
}
\vspace{-0.5cm}
\end{table}
\paragraph{Pre-training.}
We employed the hybrid architecture that incorporates both distillation and reconstruction tasks as described in Section \ref{Sec_Methodology}. For comparison with baseline models, we used the exact same pre-training data\cite{schneider2017evaluation} as in \cite{nguyen2023hyenadna, schiff2024caduceus}, which adopts the training/validation split proposed by \cite{avsec2021effective}. When pretraining on the human reference genome\cite{schneider2017evaluation}, we followed the RC equivariance inductive bias proposed by \cite{schiff2024caduceus}, implementing it using data augmentation, which has been proven to be an effective and straightforward approach.
We chose a sequence length of $1026$ for two key reasons: it's suitable for our downstream tasks (most sequences in Nucleotide Transformer Benchmarks and Genomic benchmarks are $< 1\textit{k}$ bp), and its divisibility by $6$ (the teacher's k-mer size) helps resolve tokenizer mismatches between our character-level student model and the $k$-mer based teacher model. Our student model itself is configured with $4$ Gated Delta Net (GDN) blocks, each with a dimension of $128$, resulting in a compact model with approximately $1.1$ million parameters.
Regarding the teacher model, NTv2-500M was selected as the source of high-level knowledge, providing rich feature representations.

\paragraph{Fine-tuning.}
We performed supervised training for each downstream task in both the Nucleotide Transformer benchmarks and the Genomic Benchmarks. Our fine-tuning protocol, including the use of post-hoc conjoining\cite{zhou2022towards} for model RC invariance, strictly followed the configurations outlined in \cite{schiff2024caduceus}. To ensure a fair comparison, all baselines and their reported results were adopted directly from \cite{schiff2024caduceus}, reflecting our identical experimental setup.
Evaluation metrics were chosen per benchmark: for Nucleotide Transformer Benchmarks, following \cite{nguyen2023hyenadna,schiff2024caduceus}, we used Matthews Correlation Coefficient (MCC) for histone marker tasks, F1 score for enhancer, promoter, and splice site annotation tasks (with accuracy for the ``splice site all'' task). All Genomic Benchmark tasks were evaluated using Top-1 accuracy.

\subsection{Downstream Evaluation}

\begin{table}[ht]
\centering
\caption{
    \textbf{Genomic Benchmarks Results.} Performance of HAD against baseline models. 
    Results are means from 5-fold cross-validation with 5 random seeds. The best performance in each row is in \textbf{bold}, and the second-best is \underline{underlined}.
    Error bars represent the range (maximum-minimum) across the random seeds. 
    The final row shows the average performance across all eight tasks, demonstrating HAD's strong overall results on this benchmark.
}
\label{tabs:GB_result}
\scriptsize{}
\resizebox{\linewidth}{!}{
\begin{tabular}{lccccc}
\toprule
\textbf{Dataset} & \textbf{CNN}\cite{grevsova2023genomic} & \textbf{HyenaDNA}\cite{nguyen2023hyenadna} & \textbf{Mamba}\cite{schiff2024caduceus} & \textbf{Caduceus}\cite{schiff2024caduceus} & \textbf{HAD} \\
\midrule
Mouse Enhancers           & 0.715 \rng{0.087} & \underline{0.780} \rng{0.025} & 0.743 \rng{0.054} & 0.754 \rng{0.074} & \textbf{0.788} \rng{0.033} \\
Coding vs Intergenomic    & 0.892 \rng{0.008} & 0.904 \rng{0.005}           & 0.904 \rng{0.004} & \textbf{0.915} \rng{0.003} & \underline{0.913} \rng{0.003} \\
Human vs Worm             & 0.942 \rng{0.002} & 0.964 \rng{0.002}           & 0.967 \rng{0.002} & \textbf{0.973} \rng{0.001} & \underline{0.971} \rng{0.001} \\
Human Enhancer Cohn       & 0.702 \rng{0.021} & 0.729 \rng{0.014}           & 0.732 \rng{0.029} & \textbf{0.747} \rng{0.004} & \underline{0.744} \rng{0.010} \\
Human Enhancer Ensembl    & 0.744 \rng{0.122} & 0.849 \rng{0.006}           & 0.862 \rng{0.008} & \underline{0.893} \rng{0.008} & \textbf{0.909} \rng{0.004} \\
Human Regulatory          & 0.872 \rng{0.005} & 0.869 \rng{0.012}           & 0.814 \rng{0.211} & \underline{0.872} \rng{0.011} & \textbf{0.882} \rng{0.012} \\
Human OCR Ensembl         & 0.698 \rng{0.013} & 0.783 \rng{0.007}           & 0.815 \rng{0.002} & \underline{0.828} \rng{0.006} & \textbf{0.832} \rng{0.003} \\
Human NonTATA Promoters   & 0.861 \rng{0.009} & 0.944 \rng{0.002}           & 0.933 \rng{0.007} & \underline{0.946} \rng{0.007} & \textbf{0.960} \rng{0.008} \\
\midrule
\textbf{Average}          & 0.803 & 0.853 & 0.846 & \underline{0.866} & \textbf{0.875} \\
\bottomrule
\end{tabular}
}
\vspace{-0.5cm}
\end{table}

\paragraph{Nucleotide Transformer Benchmarks.}
The evaluation of our proposed HAD model on the Nucleotide Transformer Benchmarks is presented in Table \ref{tabs:NT_result}. With only 1.1M parameters, HAD is the most compact model among all baselines, yet it demonstrates exceptional performance. It achieves leading results in the majority of Histone Marker tasks and all Enhancer Annotation tasks, securing the top position in 11 out of 18 tasks overall.  
Notably, as highlighted in the $\Delta_{\text{Student-Teacher}}$ column, HAD consistently outperforms its significantly larger teacher model (NTv2, 500M parameters) across numerous tasks, despite utilizing approximately 497.2M fewer parameters.
This outcome underscores that the feature alignment process integral to HAD not only facilitates effective knowledge transfer but also empowers the student model to surpass the teacher's performance ceiling, thereby significantly enhancing its learning and capabilities on downstream genomic tasks.
\begin{figure}[ht]
    \centering
    \includegraphics[width=0.433\linewidth]{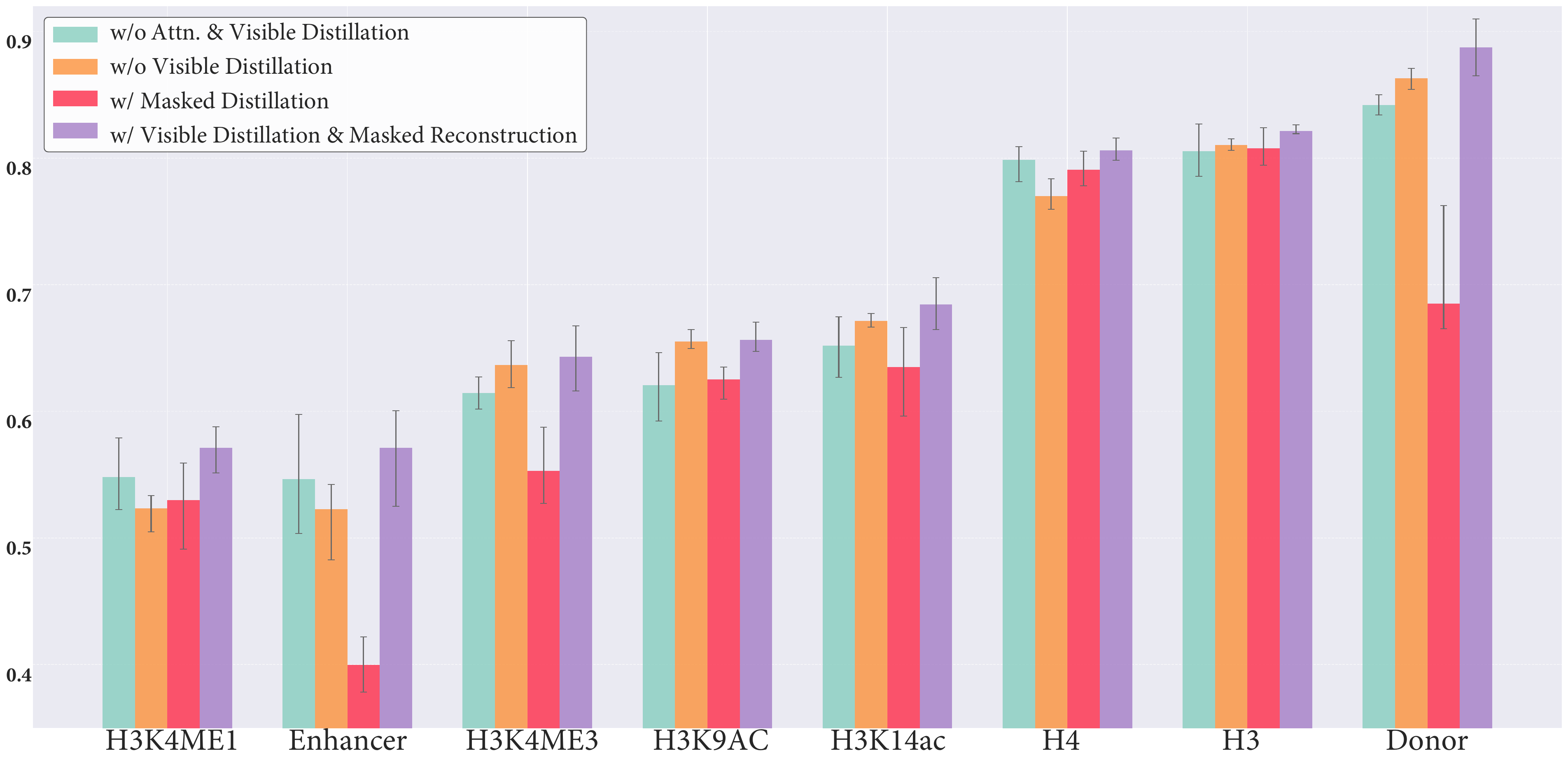}
    \includegraphics[width=0.557\linewidth]{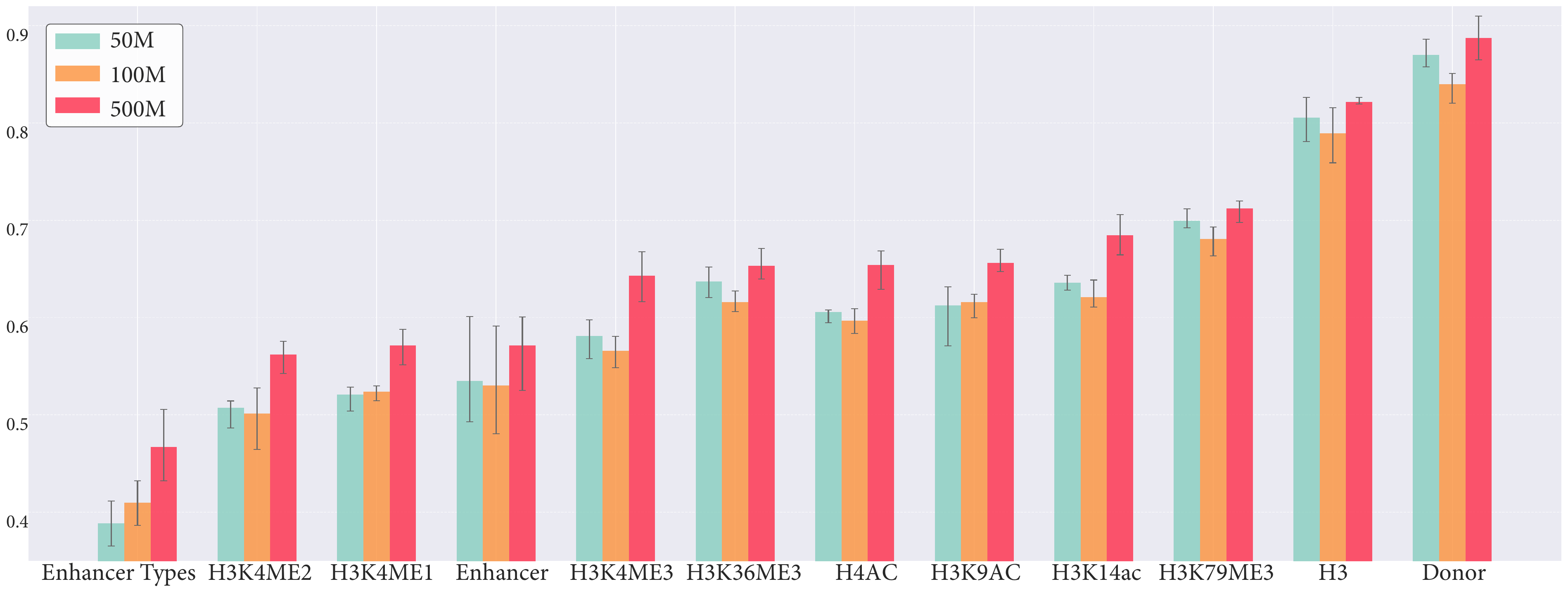}
    \vspace{-0.5cm}
    \caption{Ablation on pretraining scheme with different model architectures (left) and teacher model size (right)}
    \label{fig:ablation}
    \vspace{-0.5cm}
\end{figure}

\paragraph{Genomic Benchmarks.}
We further evaluated HAD on the Genomic Benchmarks, with results detailed in Table \ref{tabs:GB_result}. The selection of baseline models for comparison remained consistent with those in \cite{schiff2024caduceus}. In five of the eight downstream tasks within these benchmarks, our HAD model achieved the highest score among the evaluated baselines. Encouragingly, HAD's average performance score of $\textbf{0.875}$ across all tasks in these benchmarks was the highest among all reported baselines.

\subsection{Ablation Study}
\paragraph{Ablation of Different Architectures.}
\label{Sec_Ablation_Arch}
To investigate the effectiveness of our overall HAD framework, we conducted an ablation study on the Nucleotide Transformer benchmarks (Figure~\ref{fig:ablation}~left). 
Our full model (``HAD w/ Visible Distillation \& Masked Reconstruction''), with its hybrid architecture and dual-branch pretraining (visible distillation and masked reconstruction), was compared against variants lacking key components or using altered training strategies. 
These included removing self-attention and visible distillation (``HAD w/o Attn. \& Visible Distillation''; GDN backbone, MLM-only), omitting only visible distillation (``HAD w/o Visible Distillation''; hybrid architecture, MLM-only), and distilling only from masked positions (``HAD w/ Masked Distillation''). 
The full HAD model significantly outperformed these ablations, highlighting that its hybrid architecture and dual-branch learning strategy are crucial for its superior performance.

Table~\ref{tabs:ablation_arch_loss} further details the pre-training losses for these ablated architectures. 
Among the MLM-exclusive versions (``GDN w/o Attn.'' and ``GDN w/ Attn.''), both targeted masked nucleotide reconstruction; the attention-equipped version achieved lower Cross-Entropy (CE) reconstruction loss, indicating more effective pre-training. 
For distillation approaches, the MSE loss from ``$\mathcal{M}$ Dis.'' (masked distillation) was lower than the visible distillation MSE component of ``$\mathcal{V}$ Dis. \& $\mathcal{M}$ Rec.''. This discrepancy with the latter model's established superior downstream performance could be explained by two factors: first, visible distillation is more challenging due to a larger number of targets; and second, component pre-training losses often misalign with overall downstream task performance.

\paragraph{Ablation of Different Teacher Models.}
We further investigated the impact of teacher model size on distillation performance by ablating the teacher component, employing NTv2 variants with 50M, 100M, and 500M parameters. As illustrated in Figure~\ref{fig:ablation}~(right), a significant enhancement in the student model's downstream task performance was contingent upon guidance from the 500M parameter teacher. This observation highlights that a teacher model must possess substantial representational capacity, likely a product of comprehensive pretraining on extensive nucleotide data, to serve as an effective source of rich features for successful distillation to a smaller student architecture.

Further supporting this, the pre-training perplexity scores of these teacher models (Table~\ref{tabs:perplexity_results}) show a clear hierarchy: the 500M teacher achieved the lowest perplexity, followed by the 100M and 50M models across all evaluated training steps. This superior intrinsic language modeling capability of the largest teacher model likely underpins its effectiveness as a richer feature source for distillation.

\begin{table}[t]
\scriptsize
\begin{minipage}{0.49\linewidth}
    \centering
    \caption{Validation loss for different architectures.}
    \label{tabs:ablation_arch_loss} 
    \renewcommand{\arraystretch}{1}
    \setlength{\tabcolsep}{3pt}
    \begin{tabular}{lccccc} 
    \toprule
    \multirow{2}{*}{\textbf{Arch.}} & \multicolumn{5}{c}{\textbf{Step}} \\
    \cmidrule(lr){2-6}
    & $2$k & $4$k & $6$k & $8$k & $10$k \\
    \midrule
    \multicolumn{6}{l}{\texttt{Masked Language Modeling}} \\
    GDN w/o Attn. & 1.0422 & 1.0290 & 1.0202 & 1.0127 & 1.0090 \\
    GDN w/ Attn. & 1.0489 & 1.0273 & 1.0154 & 1.0068 & 1.0033 \\
    \midrule
    \multicolumn{6}{l}{\texttt{Masked Language Modeling + Distillation}} \\
    $\mathcal{M}$ Dis. & 0.3167 & 0.3149 & 0.3111 & 0.2994 & 0.2901 \\
    $\mathcal{V}$ Dis. \& $\mathcal{M}$ Rec. & 0.3177 & 0.3164 & 0.3143 & 0.3118 & 0.3049 \\
    \bottomrule
    \end{tabular}
    \vspace{-0.5cm}
\end{minipage}
\hfill
\begin{minipage}{0.55\linewidth}
    \centering
    \caption{Perplexity scores with different teacher models.}
    \label{tabs:perplexity_results}
    \renewcommand{\arraystretch}{1.15}
    \setlength{\tabcolsep}{3pt}
    \resizebox{0.8\linewidth}{!}{\begin{tabular}{cccccc}
    \toprule
    \textbf{Teacher} & \multicolumn{5}{c}{\textbf{Step}}\\
    \cmidrule(lr){2-6}
    \textbf{Model} & $2$k & $4$k & $6$k & $8$k & $10$k \\
    \midrule
    \multicolumn{6}{l}{\texttt{Masked Language Modeling + Distillation}} \\
    50M  & 5.876 & 5.835 & 5.814 & 5.797 & 5.785 \\
    100M & 5.720 & 5.679 & 5.646 & 5.622 & 5.609 \\
    500M & 5.217 & 4.903 & 4.708 & 4.599 & 4.538 \\
    \bottomrule
    \end{tabular}}
    \vspace{-0.2cm}
\end{minipage}
\end{table}
\begin{figure}[t]
    \centering
    \includegraphics[width=\linewidth]{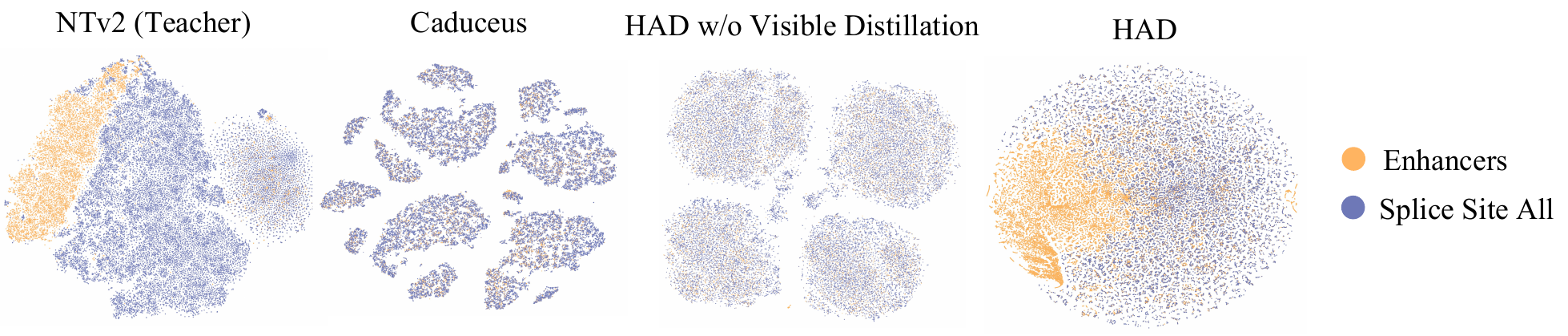}
    \vspace{-0.2cm}
    \caption{
        t-SNE visualization of pre-trained model representations on select downstream task data. 
        The clustering visually demonstrates HAD's effective knowledge transfer from its teacher NTv2 through the distillation branch, particularly for distinguishing Enhancer-related features.
    }
    \label{fig:tSNE}
    \vspace{-0.5cm}
\end{figure}

\subsection{Qualitative Analysis}
To investigate the effectiveness of feature learning specifically imparted by the distillation process within our HAD framework, we employed t-SNE \cite{van2008visualizing} to visualize representations from several models. We compared our full HAD model against the teacher model (NTv2-500M~\cite{dalla2024nucleotide}), Caduceus~\cite{schiff2024caduceus}, and the ablation variant ``HAD w/o Visible Distillation'' described in Section~\ref{Sec_Ablation_Arch}. 
These models were used as feature extractors for downstream task data from $\textit{Enhancers}$ and $\textit{splice site all}$. As visualized in Figure~\ref{fig:tSNE}, the NTv2-500M teacher model exhibits a strong capability to distinguish $\textit{Enhancers}$. In contrast, both Caduceus and the ``HAD w/o Visible Distillation'' ablation fail to form meaningful clusters for these enhancer features. Conversely, our complete HAD model clearly learns and separates these enhancer-related features, effectively mirroring its teacher's discriminative ability. This t-SNE analysis underscores that HAD successfully acquires high-level feature representations from the teacher model through the proposed distillation mechanism.
\section{Conclusion}
\label{Sec_Conclusion}

We have introduced HAD, an effective framework for DNA sequence modeling that utilize hybrid learning tasks, integrating feature alignment with masked nucleotide reconstruction.
Our compact (1.1M parameter) student model thereby learns rich, high-level biological features by distilling knowledge from an extensively pre-trained and significantly larger teacher. 
Across different downstream tasks, HAD not only outperformed most models with comparable parameters but also surprisingly exceeded the performance of its $\textbf{500}\times$ larger teacher model.
Finally, interpretable t-SNE analysis visually showcased HAD's effective knowledge transfer through its robust learning of discriminative features.
\clearpage
{
\small
\bibliographystyle{unsrt}
\bibliography{refs}
}
\end{document}